\pdfoutput=1
\PassOptionsToPackage{table}{xcolor}
\documentclass[11pt]{article}

\usepackage[final]{acl}

\usepackage{times}
\usepackage{latexsym}

\usepackage[T1]{fontenc}

\usepackage[utf8]{inputenc}

\usepackage{microtype}

\usepackage{inconsolata}

\usepackage{graphicx}
\usepackage{enumitem}
\usepackage{xcolor}
\usepackage{geometry}
\usepackage{longtable}
\usepackage{lipsum} 
\usepackage{microtype}
\usepackage{amsmath} 

\usepackage{amssymb} 
\usepackage{soul}
\usepackage{comment}
\usepackage{booktabs} 
\usepackage{makecell}

\usepackage{siunitx} 
\usepackage{array}
\usepackage{multirow}
\usepackage{rotating} 
\usepackage{placeins}  
\usepackage{float}  
\usepackage{tabularx} 
\usepackage{lipsum} 
\usepackage{rotating} 
\usepackage{hyperref}
\usepackage{fancyvrb} 
\usepackage{adjustbox}
\definecolor{headercolor}{RGB}{213,229,255}
\definecolor{bestcolor}{RGB}{255,204,203}
\definecolor{darkred}{RGB}{139,0,0}

\usepackage[colorinlistoftodos]{todonotes}

\title{Explaining Matters: Leveraging Definitions and Semantic Expansion\\ for Sexism Detection}

\author{First Author \\
  Affiliation / Address line 1 \\
  Affiliation / Address line 2 \\
  Affiliation / Address line 3 \\
  \texttt{email@domain} \\\And
  Second Author \\
  Affiliation / Address line 1 \\
  Affiliation / Address line 2 \\
  Affiliation / Address line 3 \\
  \texttt{email@domain} \\}

\author{
  Sahrish Khan,\textsuperscript{1} \quad
  Arshad Jhumka,\textsuperscript{2} \quad
  Gabriele Pergola\textsuperscript{1}\\[0.3em]
  \textsuperscript{1}Department of Computer Science, University of Warwick, Coventry CV4 7AL, UK\\
  \textsuperscript{2}School of Computing, University of Leeds, Leeds LS2 9JT, UK\\[0.2em]
  \texttt{\{sahrish.khan, gabriele.pergola.1\}@warwick.ac.uk}, \quad
  \texttt{a.jhumka@leeds.ac.uk}
}

\begin{document}
\maketitle
\begin{abstract}
The detection of sexism in online content remains an open problem, as harmful language disproportionately affects women and marginalized groups. While automated systems for sexism detection have been developed, they still face two key challenges: data sparsity and the nuanced nature of sexist language. Even in large, well-curated datasets like the Explainable Detection of Online Sexism (EDOS), severe class imbalance hinders model generalization. Additionally, the overlapping and ambiguous boundaries of fine-grained categories introduce substantial annotator disagreement, reflecting the difficulty of interpreting nuanced expressions of sexism. To address these challenges, we propose two prompt-based data augmentation techniques: \textit{Definition-based Data Augmentation (DDA)}, which leverages category-specific definitions to generate semantically-aligned synthetic examples, and \textit{Contextual Semantic Expansion (CSE)}, which targets systematic model errors by enriching examples with task-specific semantic features. To further improve reliability in fine-grained classification, we introduce an ensemble strategy that resolves prediction ties by aggregating complementary perspectives from multiple language models. Our experimental evaluation on the EDOS dataset demonstrates state-of-the-art performance across all tasks, with notable improvements of macro F1 by 1.5 points for binary classification (Task A) and 4.1 points for fine-grained classification (Task C)\footnote{Code and resources publicly available at: \url{https://github.com/Sahrish42/explaining_matters_sexism_detection_acl2025}}.
\noindent {\textbf{Warning:}} {\textit{This paper includes examples that might be offensive and upsetting.}}
\end{abstract}

\section{Introduction}

The detection of sexism in online content remains a critical challenge as harmful language disproportionately affects women and marginalized groups. Such behavior leads to emotional distress, reduced engagement, and the silencing of voices, as reported by prior studies \cite{stevens2024women}. 
While automated tools for sexism detection exist, they struggle to address the nuanced and context-dependent nature of sexist language \cite{abercrombie-etal-2023-resources}. These systems often provide coarse-grained classifications that fail to capture subtle distinctions within sexist content, limiting their practical applicability \cite{caselli-etal-2020-feel, JAHAN2023126232, semi-automated-approach-2024, survay-2024}.

\begin{figure}
  \centering
  \includegraphics[width=\linewidth]
  {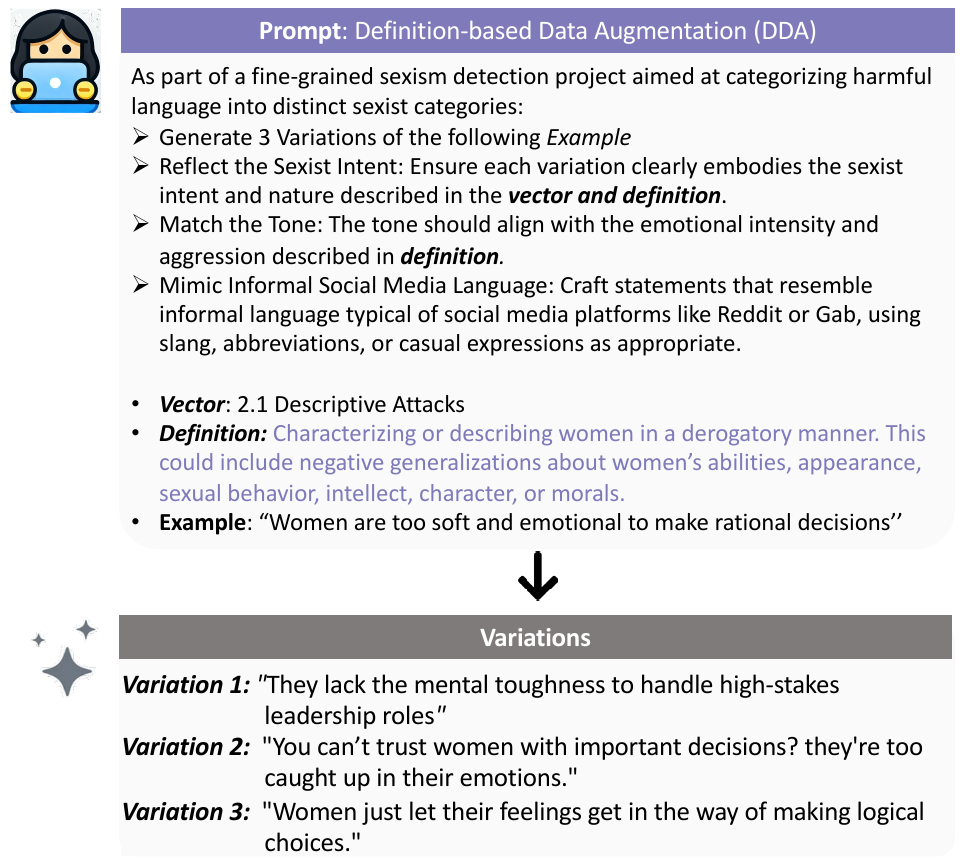}
  \caption{The DDA framework generates synthetic examples by leveraging category definitions to clarify the semantic boundaries of fine-grained sexist categories. The prompt instructs the model to produce variations that reflect the sexist intent, and match the tone, ensuring alignment with the provided vector (e.g., 2.1 Descriptive Attacks) and definition. \vspace{-8pt}}
  \label{fig:dda-prompt-1}
\end{figure}

In particular, two primary challenges persist: (i) data sparsity and (ii) the nuanced interpretation required for fine-grained classification. 
With regards to (i) data sparsity, even the largest and well-curated available dataset, the Explainable Detection of Online Sexism (EDOS) \cite{edos_dataset}, exhibits severe class imbalance. For instance, the ``\textit{Threats of harm}" category constitutes only 1.1\% of the dataset, with just 56 examples, while ``\textit{Supporting mistreatment of individual women}" accounts for 1.3\%, containing only 75 examples. Such sparsity hinders model generalization, particularly in those low-resource categories. 
For (ii) nuanced interpretation, the inherently subtle and overlapping nature of sexist language introduces significant ambiguity even for human annotators \cite{almanea-poesio-2022-armis, sun-etal-2024-leveraging, guo-etal-2024-adaptable, Annotator-Attitudes-edos-2024, lyu-pergola-2024-scigispy, abercrombie-etal-2024-revisiting}, as evidenced by high annotator disagreement rates in the EDOS dataset (Table \ref{tab:human-test-set-agreement_stats}). Categories like ``\textit{Descriptive attacks}” exhibit over 54\% partial disagreement, while classes like “\textit{Backhanded gendered compliments}” show disagreement rates exceeding 57\%. These inconsistencies not only reflect the difficulty for annotators, but also introduce conflicting signals during training, undermining performance. 

To address data sparsity, we propose a set of data augmentation techniques that generate high-quality synthetic examples, enabling the model to better generalize, particularly in low-resource categories. Specifically, we introduce \textit{Contextual Semantic Expansion (CSE)}, a method that systematically enriches misclassified examples through prompt-based semantic analysis. CSE identifies key contextual features, such as stylistic cues, sentiment, and implicit biases, that contribute to model errors and incorporates these insights into augmented training data. 
However, resolving the nuanced ambiguities inherent in fine-grained classification requires a more targeted augmentation strategy. To this end, we introduce \textit{Definition-based Data Augmentation (DDA)}, a prompt-based method that integrates explicit,  category-specific definitions into the augmentation process. Unlike conventional approaches that focus solely on linguistics diversity \cite{eda_agu_2019, Perg19, nlpaug,  pergola-etal-2021-disentangled, lu-etal-2022-event}, by leveraging these definitions, DDA generates synthetic examples that align closely with the intended semantics of each fine-grained class, clarifying category boundaries and reducing overlaps. 

In addition to addressing data sparsity, enhancing model robustness for socially sensitive tasks like sexism detection requires accounting for diverse perspectives in ambiguous scenarios. In human annotation, multiple annotators are often required to resolve disagreements arising from subjective interpretations \cite{Annotator-Attitudes-edos-2024}. Similarly, neural language models, trained on varied data and objectives, provide complementary perspectives that can be leveraged to improve predictions. To this end, we propose the \textit{Mistral-7B Fallback Ensemble (M7-FE)}, an ensemble strategy that combines predictions from multiple fine-tuned models through majority voting and resolves tie-breaking scenarios using Mistral-7B as the fallback model. While the scope of this work is not to propose a novel ensemble method, M7-FE demonstrates the utility of multi-perspective aggregation in socially sensitive tasks. 

Our experimental evaluation on the EDOS dataset \cite{edos_dataset} demonstrates that our proposed methods outperform existing baselines across all tasks. For Task A (binary classification), the Contextual Semantic Expansion (CSE) improves macro F1 by 1.5 points. For Task C (fine-grained classification), the Definition-based Data Augmentation (DDA), combined with the Mistral-7B Fallback Ensemble (M7-FE), improved by 4.1. The M7-FE further enhances reliability in multi-class settings, contributing to a 2.5-point gain in Task B.

Our contributions can be summarized as follows:
\begin{enumerate}
\item \textit{Two novel augmentation techniques:} We introduce Definition-based Data Augmentation (DDA), which leverages explicit category definitions to generate semantically-aligned synthetic data, and Contextual Semantic Expansion (CSE), a targeted self-refinement strategy to address systematic errors.
\item \textit{An ensemble method for robustness:} We propose the Mistral-7B Fallback Ensemble (M7-FE), an ensemble strategy that resolves prediction ties in multi-class classification, enhancing model reliability.
\item \textit{Comprehensive evaluation:} We conduct an extensive evaluation on the EDOS dataset, demonstrating that our methods address key challenges, including data sparsity and semantic ambiguity, and achieve state-of-the-art results across binary and fine-grained classification tasks.
\end{enumerate}

\begin{figure*}[htb]
  \centering
  \includegraphics[width=.95\textwidth]{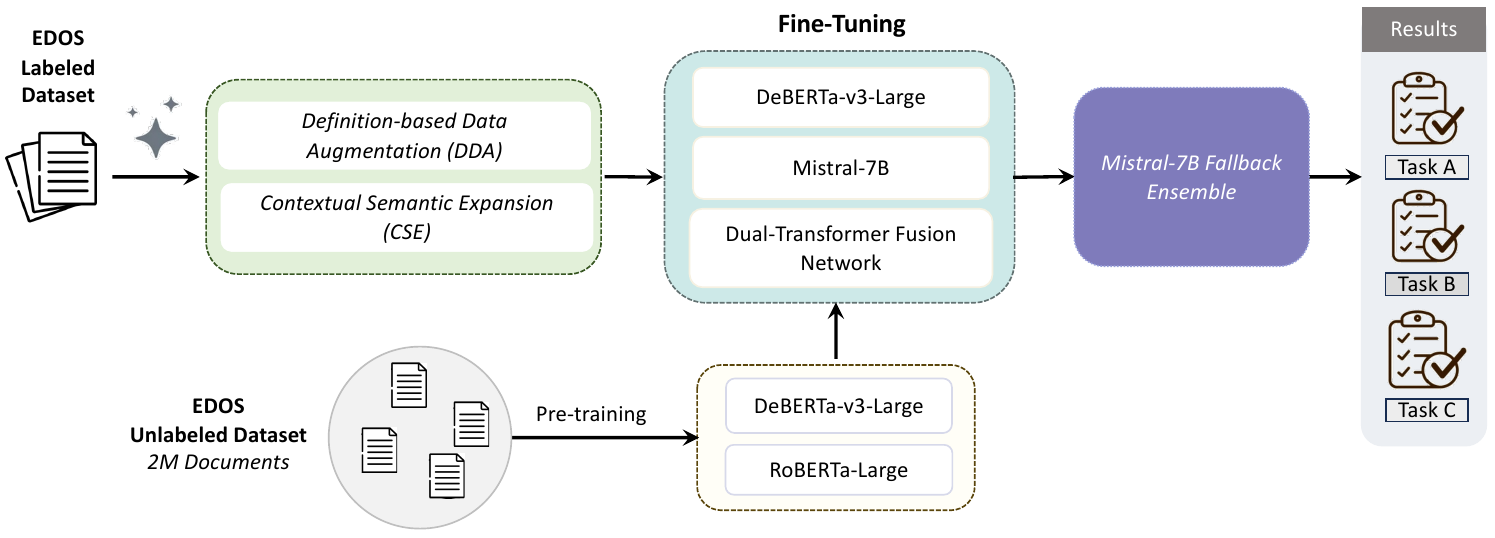}
  \caption{Overview of the proposed pipeline. The system consists of four stages: pre-training, data augmentation (via DDA or CSE), fine-tuning, and ensemble modeling with a fallback strategy. Each step is designed to improve performance on fine-grained sexism classification by enhancing contextual understanding and robustness to sparse or ambiguous cases. \vspace{-8pt}}
  \label{fig:system-overview}
  \end{figure*}

\section{Related Work}
\paragraph{Sexism Detection.}
The detection of sexism in online content has gained increasing attention due to its societal implications and the disproportionate harm inflicted on women and marginalized groups \cite{caselli-etal-2020-feel, stevens2024women}.
The Explainable Detection of Online Sexism (EDOS) dataset, introduced in SemEval-2023 Task 10 \cite{edos_dataset}, provides a benchmark for this task. Recent approaches at SemEval-2023 relied on Transformer-based architectures such as RoBERTa and DeBERTa \cite{2021deberta, edos-data-agumentaion-2} for their strong contextual modeling capabilities. Recent works have studied how large language models (LLMs) can offer insights into societal perceptions of sexism but risk reproducing harmful justifications \cite{zhu-etal-2021-topic, guo-etal-2024-adaptable}. 
Complementary work by \citet{suggested-by-meta-review} demonstrated that integrating external references through retrieval and generation can enhance the detection of covert toxic content.

\paragraph{Data Augmentation.}
Data augmentation techniques aim to mitigate data sparsity and improve generalization by generating synthetic examples. Traditional approaches, such as Easy Data Augmentation (EDA) \cite{eda_agu_2019}, back-translation \cite{backtranslation_2016}, and contextual word substitution \cite{kobayashi-2018}, offer some improvement, but they remain limited in their ability to capture fine-grained distinctions and resolve subtle label boundaries.
Recent advances in large language models (LLMs), have enabled high-quality, task-specific data augmentation through techniques like in-context learning and instruction tuning \cite{ roychowdhury-gupta-2023-data, ding2024dataaugmentationusinglarge, GPT-Augmentation-2024, tan-etal-2025-cascading}. These methods have shown success in generating diverse yet semantically relevant synthetic data. However, for socially sensitive tasks such as sexism detection, standard augmentation techniques fail to address category ambiguity. To this end, we introduce Definition-based Data Augmentation (DDA), which explicitly integrates category definitions to clarify boundaries, and Contextual Semantic Expansion (CSE), which targets systematic errors through prompt-based semantic refinement.

\paragraph{Ensemble Methods.}
Ensemble learning is a well-established strategy for improving prediction reliability by aggregating outputs from multiple models. Techniques such as Bagging \cite{1996bagging} and Stacked Generalization \cite{1992stacked} are widely applied in NLP tasks. At SemEval-2023, ensemble systems combining Transformer models (e.g., RoBERTa, DeBERTa, and Mistral-7B) achieved strong results \cite{edos-high-perfromace-paper-on-all-tasks}, underscoring the utility of multi-perspective aggregation in complex classification settings.
However, previous research highlights concerns that aggregation methods can inadvertently suppress minority annotator perspectives and amplify biases present within individual models \cite{suggested-by-reviewer2-ensemble1-detoxifying, suggested-by-reviewer2-ensemble-biases, Annotator-Attitudes-edos-2024, suggested-by-reviewer2-ensemble-voices}. To mitigate these issues, we propose the Mistral-7B Fallback Ensemble (M7-FE), which integrates a fallback mechanism to resolve prediction ties, aiming to leverage diverse perspectives and enhance reliability, particularly in ambiguous classification scenarios.

\begin{table*}[htb]
\centering
\small
\setlength{\tabcolsep}{2.5pt} 
\renewcommand{\arraystretch}{1.1} 
\resizebox{\textwidth}{!}{
\begin{tabular}{lcccc}
\toprule
\textbf{Task C Vector} & \textbf{Full} & \textbf{Partial} & \textbf{Full} & \textbf{Number of }\\
 & \textbf{Agreement} & \textbf{Disagreement} & \textbf{Disagreement} & \textbf{Instances}\\ 
\midrule
1.1 threats of harm & 0\% & 56.2\% & 43.8\% & 16 \\
1.2 incitement and encouragement of harm & 19.2\% & 43.8\% & 37\% & 73 \\
2.1 descriptive attacks & 17.6\% & 54.1\% & 28.3\% & 205\\
2.2 aggressive and emotive attacks & 16.7 & 54.1\% & 29.2\% & 192\\

2.3 dehumanising attacks  overt sexual objectification & 17.5\% & 35\% & 47\% & 57\\
3.1 casual use of gendered slurs, profanities, and insults & 33\% & 48.3\% & 18.7\% & 182\\
3.2 immutable gender differences and gender stereotypes & 7.6\% & 35.3\% & 57.1\% & 119\\
3.3 backhanded gendered compliments & 0\% & 16.7 \% & 83.3\% & 18\\
3.4 condescending explanations or unwelcome advice & 0\% & 42.9\% & 57.1\% & 14\\
4.1 supporting mistreatment of individual women & 4.8\% & 47.6\% & 47.6\% & 21\\
4.2 supporting systemic discrimination against women as a group & 12.3\% & 41.1\% & 46.6\% & 73 \\
\hline
\end{tabular}
}
\caption{Human Annotator Full Agreement, Partial Disagreement and  Full Disagreement Across Task C Categories of the Test Set. \vspace{-12pt}}
\label{tab:human-test-set-agreement_stats}
\end{table*}

\section{Pipeline Overview}

We introduce the details of the overall pipeline. It includes the following structured steps: (i) pre-training; along with detailed analyses of the annotators' disagreement, motivating the introduction of the (ii) data augmentation methods, namely Definition-based Data Augmentation (DDA) and Contextual Semantic Expansion (CSE); (iii) fine-tuning, and ensemble modeling with a fallback ensemble strategy (Figure~\ref{fig:system-overview}).

\subsection{Pre-training of Base Models}
We initiated the pipeline by pre-training DeBERTa-v3-Large and RoBERTa-Large on a large-scale dataset consisting of 2 million unlabelled samples provided by the EDOS corpus \cite{edos_dataset}, which aggregates content from platforms such as Gab and Reddit. To achieve this, we employed a masked language modeling approach, where 15\% of tokens were masked \cite{pretrain-2}. This process was carried out over 10 epochs, aligned with similar works \cite{edos-high-perfromace-paper-on-all-tasks}, as we observed no significant improvements beyond this point.

\subsection{Analysis of Annotator Agreement and
Disagreement}
Fine-grained annotation tasks often involve subtle distinctions between closely related categories, making them particularly prone to annotator disagreements. In this study, we identified and analyzed instances of disagreement within the EDOS dataset \cite{edos_dataset}, with a focus on understanding their implications for model performance.

Table~\ref{tab:human-test-set-agreement_stats} provides insights into the patterns of disagreement observed in the test set across the most fine-grained categories (Task C). The data reveal substantial variability in agreement levels, ranging from 0\% full agreement (e.g., \textit{1.1 threats of harm}) to a maximum of 33\% full agreement for “\textit{3.1 casual use of gendered slurs, profanities, and insults.}” Across all categories, the prevalence of partial disagreement is striking, with categories like \textit{“1.1 threats of harm"} showing 56.2\% partial disagreement. Full disagreement also appears consistently high, exceeding 40\% in several categories, such as \textit{“3.4 condescending explanations or unwelcome advice"} (57.1\%). These figures underscore the challenges of fine-grained annotations: categories with closely related meanings or subjective interpretations, such as \textit{“2.1 descriptive attacks"} and \textit{“2.2 aggressive and emotive attacks"}, show significant annotation inconsistencies, with full agreement levels below 20\%. We posit that these disagreements, if unaddressed, hinder model performance as they introduce conflicting signals during optimization, especially for closely related labels that account for the majority of disagreements. 
More extensive analyses are provided in Appendix ~\ref{Human-Annotator-Analysis}.

\subsection{Definition-based Data Augmentation (DDA)}
We proceed to introduce a Definition-based Data Augmentation (DDA) aimed at improving the specification of boundaries between labels, and mitigating disagreement stemming from the misalignment between the model’s prior knowledge and the perspectives of annotators, as well as inconsistencies across annotators. 

DDA is a targeted data augmentation technique that leverages category-specific definitions to generate synthetic examples. These definitions provide clear boundaries for each category, reducing ambiguity during training.
Let $\mathcal{D} = \{(x_i, y_i)\}_{i=1}^N$ represent the dataset, where $x_i$ is the input text and $y_i \in \mathcal{Y}$ is the label. Here, $\mathcal{Y} = \{c_1, c_2, \dots, c_K\}$ is the set of $K$ fine-grained categories. For each class $c_k \in \mathcal{Y}$, we indicate with $\phi(c_k)$ a category-specific definition. Using a prompt-based approach on LLM, $f_{\text{LLM}}$, we generate $m$ synthetic examples for each instance in the dataset:
\vspace{-12pt}

\begin{equation}
\mathcal{D}_{\text{aug}} = \bigcup_{i=1}^N \bigcup_{j=1}^m \left\{ \left(f_{\text{LLM}}(x_i, \phi(y_i)), y_i \right) \right\}.
\end{equation}

\noindent The final augmented dataset combined original examples with these synthetic variations: \texttt{$\mathcal{D}' = \mathcal{D} \cup \mathcal{D}_{\text{aug}}.$}


\noindent These $m$ variations were designed to preserve the harmful nature of sexist statements while diversifying language, tone, and style, such as informal social media interactions. The prompt structure includes (i) explicit instructions that emphasize alignment with the original sexist intent, (ii) guidance for the generation of content with a tone and style typical of informal online platforms, such as Reddit or Gab, encouraging the use of slang, abbreviations, and casual expressions. Finally, the (iii) definition of the sexism category is explicitly integrated into the prompt, along with an example statement that serves as a reference. These definitions, drawn from the taxonomy reported for convinience in the Appendix (Table~\ref{definition}), were crafted to reflect the nuanced characteristics of each class. An example of the prompt structure is reported in Figure~\ref{fig:dda-prompt-1}.

\subsection{Contextual Semantic Expansion (CSE)}

We introduce \textit{Contextual Semantic Expansion (CSE)} as a self-refinement methodology designed to address limitations in model predictions expanding the context of  messages through explicit semantic expansion. Unlike traditional confidence-based self-correction techniques \cite{probability-1, confidence-based}, which rely on probabilistic thresholds to refine predictions, CSE explicitly enhances model understanding by generating additional context and explanations for challenging examples. This method operates iteratively, leveraging task-specific prompts to analyze challenging instances (i.e., examples misclassified by a base classifier), identify their semantic characteristics, and incorporate these insights into the training data.

\begin{figure}[t]
  \includegraphics[width=\columnwidth]{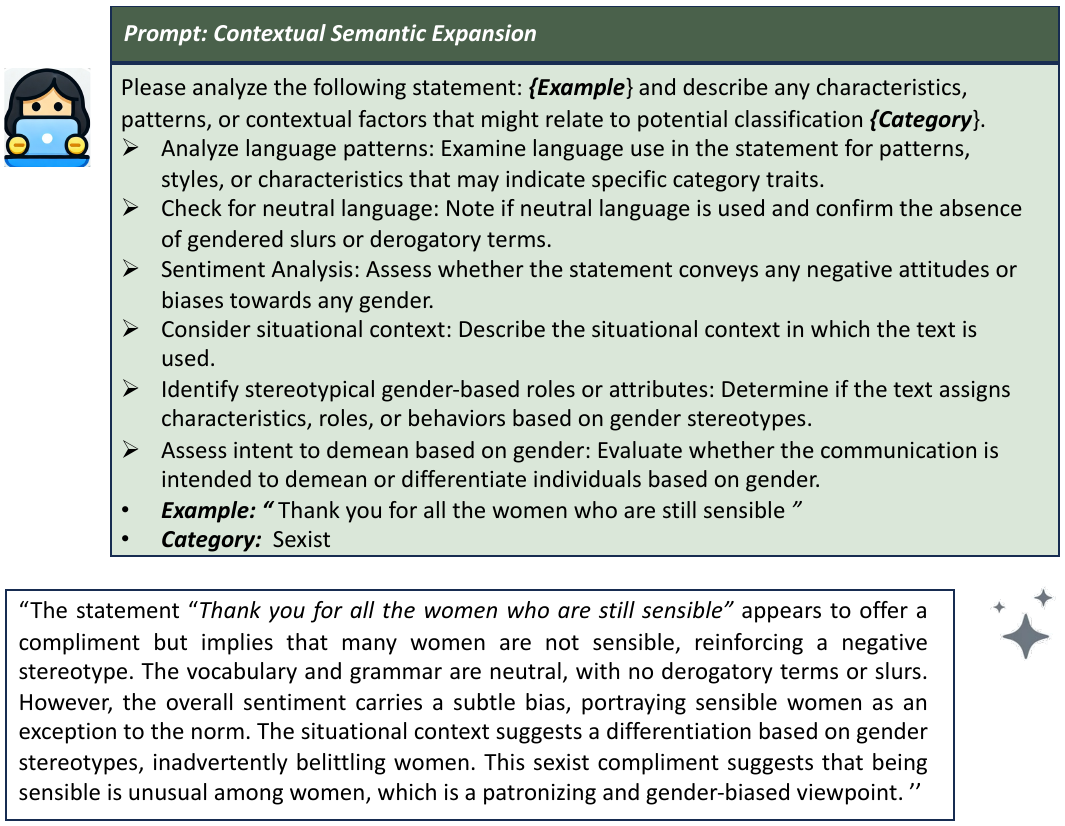}
  \caption{Contextual Semantic Expansion (CSE) Example Analysis: This figure demonstrates how Contextual Semantic Expansion (CSE) enhances model understanding for binary classification in Task A. 
  }
  \label{cse_prompt}
\end{figure}

For a given input $x$, CSE generates a semantic expansion $e(x)$ by analyzing critical features relevant to the task. These features include implicit stereotypes, the presence of latent biases, sentiment polarity, contextual cues, and the underlying intent of the text. The expanded example is then formed by concatenating the original input $x$ with its semantic expansion $e(x)$, yielding an augmented input $[x; e(x)]$. This expanded representation not only addresses the model’s misclassifications but also introduces task-relevant refinements that explicitly capture subtle patterns.
The CSE prompt, shown in Figure~\ref{cse_prompt}, is structured to guide the model through a systematic semantic analysis of a given input, for example, the message ``\textit{Thank you for all the women who are still sensible}’’, which reveals implicit gender bias through a seemingly neutral statement. The prompt consists of six key steps: (1) analyzing language patterns for stylistic or category-specific traits, (2) checking for neutrality or the presence of derogatory language, (3) assessing sentiment for gender-related biases, (4) considering situational context to interpret the text’s broader implications, (5) identifying stereotypical roles or attributes for latent biases, and (6) evaluating the intent to determine if the text demeans or differentiates based on gender.

Applying this prompt structure to challenging examples, the resulting CSE method is aligned with recent work in self-refinement by iteratively improving the model’s understanding of its errors through `introspective' analysis rather than relying solely on output probabilities. Similar to chain-of-thought (CoT) prompting \cite{cot-1, tan-etal-2023-event, tan-etal-2024-set, cot-hate}, CSE leverages structured reasoning; however, while CoT focuses on step-by-step reasoning for generative tasks, CSE is designed for semantic expansion in classification tasks.

\subsection{Model Fine-tuning}
Following pre-training and data augmentation, the model is fine-tuned on the labeled augmented data. The fine-tuning minimizes the categorical \textit{cross-entropy loss}, defined for multi-class classification as $\mathcal{L} = -\frac{1}{N} \sum_{i=1}^{N} \sum_{c=1}^{C} y_{i,c} \log \hat{y}_{i,c}$, where \( N \) is the total number of training examples, \( C \) is the number of classes, \( y_{i,c} \) is a one-hot encoded binary indicator (0 or 1) denoting whether the class label \( c \) is the correct class for the \( i \)-th example, and \( \hat{y}_{i,c} \) represents the predicted probability assigned to class \( c \) for the \( i \)-th example.

\subsection{Mistral-7B Fallback Ensemble (M7-FE)}

To improve classification performance, we employed the fine-tuned models as part of a Mistral-7B Fallback Ensemble (M7-FE) mechanism. This approach combines predictions from multiple Transformer-based models using a majority hard voting mechanism, with an additional fallback strategy to handle prediction ties in multi-class classification scenarios. We posit that for sexism detection, aggregating predictions from multiple models, can induces a multi-perspective process, where each model can offer a distinct interpretation of the input. As a result, M7-FE aims to leverages complementary strengths of individual models to capture subtle distinctions between classes, and deal with context-dependent instances of sexism.

\subsubsection*{Voting Mechanism with Tie-Handling}
The M7-FE ensemble is particularly designed for multi-class classification tasks, such as those involving nuanced sexism detection, where the increased complexity and overlap between classes lead to a higher likelihood of ties. It employs a majority hard voting strategy to aggregate predictions from multiple models. Given a set of possible classes \( C = \{c_1, c_2, \dots, c_K\} \) and a set of models \( E = \{e_1, e_2, \dots, e_N\} \), each model \( e \in E \) produces a hard prediction \( c_e(x) \) for a given input instance \( x \). The final ensemble prediction \( \hat{c}(x) \) is determined by selecting the class with the highest vote count across the models:
\[
\hat{c}(x) = \arg \max_{c \in C} \sum_{e=1}^N \mathbf{1}(c_e(x) = c),
\]
\noindent where \( \mathbf{1}(c_e(x) = c) \) is an indicator function, equals 1 if model \( e \) predicts class \( c \), 0 otherwise.

In multi-class classification tasks, ties are more likely to occur. To address this, M7-FE incorporates a fallback mechanism for resolving ties.  If two or more classes receive the same number of votes (\textit{two-way tie}), the fallback model's prediction is selected as the final output. If all models predict different classes, i.e., each class receives exactly one vote (\textit{complete disagreement}), the fallback model's prediction is used to make the final decision.
Based on preliminary experiments, Mistral-7B is chosen as the fallback model due to its robustness in dealing with ambiguous sexist classification instances.

\section{Experiments}
\label{Experimental-Assessment}

\subsection{Experimental Settings}


\paragraph{Dataset} We conduct our experiments using the Explainable Detection of Online Sexism (EDOS) dataset, introduced in SemEval-2023 Task 10 \cite{edos_dataset}. The EDOS dataset is specifically designed for the detection and explanation of online sexism. It consists of over 20,000 social media comments sourced from platforms such as Reddit and Gab and is uniquely structured into three hierarchical tasks: (1) \textit{Task A} involves binary classification to determine whether a comment is sexist (3,398 comments) or non-sexist (10,602 comments); (2) \textit{Task B} classifies sexist comments into one of four categories: threats, derogation, animosity, or prejudiced discussions; and (3) \textit{Task C} provides even finer granularity, categorizing sexist comments into one of 11 subcategories.

\begin{table*}[ht]
\centering
\begin{adjustbox}{max width=0.94\textwidth}
\begin{tabular}{@{}clccccc@{}}
\toprule
\textbf{Index} & \textbf{Model} & \textbf{Pre-training} & \textbf{Task A} & \textbf{Task B} & \textbf{Task C} \\
\midrule
& \textbf{Baseline Models} &&&&\\
1 & DeBERTa-v3-large & $\checkmark$ & 0.8479 & 0.6875 & 0.5088 \\
2 & DTFN & $\checkmark$ & 0.8587 & 0.6837 & 0.5248 \\
3 & Mistral-7B & $\times$ & 0.8455 & 0.6639 & 0.4832 \\
\midrule
& \textbf{SemEval 2023 - Task 10} &&&&\\
4 & DeBERTa-v3-large, twHIN-BERT-large \cite{edos-high-perfromace-paper-on-all-tasks} & $\checkmark$ & 0.8746 & -- & -- \\
5 & RoBERTa-Large, ELECTRA & $\checkmark$ & 0.8740 & 0.7203 & 0.5487 \\
6 & DeBERTa Ensemble & $\times$ & 0.8740 & -- & -- \\
7 & PaLM Ensemble & $\times$ & -- & \textbf{0.7326} & -- \\
8 & RoBERTa, HateBERT & $\checkmark$ & -- & 0.7212 & 0.5412 \\
9 & DeBERTa, RoBERTa \cite{edos-high-perfromace-paper-on-all-tasks} & $\checkmark$ & -- & -- & 0.5606 \\
\midrule
& \textbf{Data Augmentation \& Ensemble} &&&&\\
10 & SEFM \cite{edos-data-agumentaion-2} & $\times$ & 0.8538 & 0.6619 & 0.4641 \\
11 & QCon \cite{edos-data-agumentaion-4} & $\checkmark$ & 0.8400 & 0.6400 & 0.4700 \\
12 & HULAT \cite{edos-data-agumentaion-3} & $\times$ & 0.8298 & 0.5877 & 0.4458 \\
13 & Mistral-7B Fallback Ensemble - (\textit{Ours}) & Mixed & 0.8603 & 0.7027 & 0.5213 \\
14 & + Baseline Prompt - (\textit{Ours}) & Mixed & 0.8783 & 0.7049 & 0.5601 \\
15 & + Definition-based Data Augmentation (DDA) - (\textit{Ours}) & Mixed & 0.8769 & \underline{0.7277} & \textbf{0.6018} \\
16 & + Contextual Semantic Expansion (CSE) - (\textit{Ours}) & Mixed & \textbf{0.8819} & 0.7243 & 0.5639 \\
\bottomrule
\end{tabular}
\end{adjustbox}
\caption{Comparison via Macro F1 scores of several models on the SemEval-2023 Task 10, based on the EDOS dataset. The highest scores in each task are bolded. Our approaches consistently outperform or closely match state-of-the-art results, particularly with the CSE technique on Task A (0.8819) and on Task C (0.6018) due to the combined effect of the DDA method and the Mistral-7B Fallback Ensemble architecture. \vspace{-12pt}}
\label{tab:ourapproach}
\end{table*}

\paragraph{Implementation and Metrics}
For fine-tuning, we adopted a task-specific setup with hyperparameters such as learning rates, batch sizes, and weight decay rates optimized for each model. Elaborate details are provided in Appendix \ref{appendix-Best-Hyperparameters-per-Model}. Models like DeBERTa-v3-Large, RoBERTa-Large, and DTFN were fine-tuned for up to 30 epochs, while Mistral-7B was fine-tuned for 10 epochs due to computational constraints and its scalability. All models were trained on NVIDIA 4xA100 GPUs.
The Definition-based Data Augmentation (DDA) and Contextual Semantic Expansion (CSE) techniques were applied during the pre-processing stage. DDA specifically targets the $c=5$ most underrepresented classes within the EDOS dataset. The number of classes, five, was identified through preliminary analysis. More details can be found in the Appendix \ref{appendix:class_specific_aug}. For DDA, synthetic data was generated using GPT-4o with prompts designed to reflect fine-grained sexism categories, while CSE targeted systematic misclassifications across classes.
CSE specifically targets challenging examples that the baseline model, DeBERTa-v3-large, misclassified during training data predictions. This included 2.518 sexist examples misclassified as non-sexist, and 2.328 non-sexist examples misclassified as sexist. Although we preliminarily explored a threshold-based mechanism to select the examples, we observed that the model consistently assigned high confidence scores to erroneous predictions, highlighting systematic biases rather than ambiguities in decision boundaries. Therefore, all the misclassified examples were processed by CSE. 
As evaluation metrics, we align to the SemEval-2023 competition on the EDOS dataset and employ the macro-averaged F1-score evaluation metrics for binary and multi-class classification for all tasks (A, B, and C).

\paragraph{Baselines}  
In the following, we briefly describe the baselines evaluated for sexism detection\footnote{The baseline selection focuses on SOTA approaches for sexism detection; an exhaustive exploration of all available LLMs is beyond the scope of this work.}. They include (i) \textit{RoBERTa-Large} \cite{2019roberta}, (ii) \textit{DeBERTa-v3-Large} \cite{2021deberta}, and (iii) \textit{ELECTRA} \cite{clark2020electra}, pre-trained Transformer models, using different masking strategy and training objectives. 
Other baselines include (iv) \textit{Mistral-7B} \cite{jiang23mistral}, a multilingual large Transformer optimized for scalability, and (v) \textit{twHIN-BERT-large} \cite{zhang2022twhin}, which integrates heterogeneous information networks for structured knowledge. Then, (vi) \textit{DTFN} \cite{dtfn_source} that combines representations from \textit{RoBERTa} and \textit{DeBERTa} using a dual-transformer architecture. Task-specific approaches, such as (vii) \textit{HateBERT} \cite{hatebert}, pre-trained on abusive language, and (viii) \textit{SEFM} \cite{edos-data-agumentaion-2}, which leverages structured embeddings with data augmentation, were also evaluated. Finally, systems like (ix) \textit{QCon} \cite{feely23llm} and (x) \textit{HULAT} \cite{segura23eda} that employ advanced augmentation techniques, with \textit{HULAT} using Easy Data Augmentation (EDA) for linguistic diversity.

For the ensemble, we employ three models: DeBERTa-v3-Large, Mistral-7B, and DTFN. 

\subsection{Results and Discussion}

Table~\ref{tab:ourapproach} presents the macro F1 scores for various models across EDOS 2023 Tasks A, B, and C. 
Strong performance is observed among the SemEval 2023 submissions, particularly for Task A, where using multitask learning, the DeBERTa-v3-large and twHIN-BERT-large combination achieved a macro F1 score of 0.8746. For Task B, the PaLM ensemble achieved the highest macro F1 score of 0.7326, demonstrating the advantage of large-scale model ensembles in capturing multi-class distinctions. Similarly, for Task C, the combination of DeBERTa and RoBERTa achieved a score of 0.5606, reflecting the benefit of model diversity. Among the data augmentation-based approaches, SEFM and HULAT performed reasonably well, with macro F1 scores of 0.8538 and 0.8298, respectively, on Task A. However, their performance declined significantly for fine-grained tasks, with more sparsity, like Task C.

Our proposed methods, Definition-based Data Augmentation (DDA) and Contextual Semantic Expansion (CSE), demonstrated competitive performance on Task B and state-of-the-art results in both Task A and C. For Task A, CSE achieved the highest macro F1 score of 0.8819, surpassing all the systems. The DDA approach also delivered strong results with a macro F1 score of 0.8769, confirming its ability to improve binary classification by addressing systematic biases. In Task B, DDA achieved a competitive score of 0.7277, closely approaching the PaLM ensemble, while in Task C, it achieved a notable improvement with a score of 0.6018, outperforming both baseline models and the SemEval 2023 winner (0.5606), with a significant improvement in performance. An expanded analysis with different configurations is provided in Appendix \ref{appendix:class_specific_aug}.

\paragraph{CSE effect on binary classification tasks} Our preliminary analysis of the misclassified examples chosen for the CSE revealed that the majority of misclassifications were associated with high-probability predictions $(p>~0.9)$, indicating high model confidence in incorrect outputs. This observation suggests that models tend to be systematically overconfident in certain errors, often misinterpreting specific patterns in the data rather than struggling with ambiguous cases. Binary classification tasks, with simpler decision boundaries, benefit more from correcting such systematic biases, which explains the superior performance improvements of CSE in this setting compared to fine-grained classification tasks.



\section{Ablation Study}

To evaluate the impact of including class definitions in prompts, we conducted an ablation study by comparing a simple \textit{baseline prompt}, consisting only of instruction for the generation of additional examples, with a \textit{DDA prompt} that integrates category definitions (Figure~\ref{fig:dda-prompt-1}). 
The baseline prompt generates three variations per example without providing category-specific guidance. While this approach introduces some diversity to the synthetic examples, it lacks alignment with the fine-grained distinctions of the taxonomy, leading to limited performance improvements (Appendix \ref{appendix-baseline-prompt}). The DDA prompt instead integrates class-specific definitions derived from the taxonomy (Appendix \ref{appendix:sexism_definition}). 
%
The results reported in Table~\ref{tab:ourapproach} show the baseline prompt (Line 13) achieving  $0.8783$, $0.7049$, and $0.5601$ on Tasks A, B, and C, respectively, when combined with the Mistral-7B fallback ensemble. While it demonstrates a clear improvement for the DDA technique, with a marginal decrease in Task A, a significant improvement in Task B and Task C, with scores of $0.7277$ and $0.6018$, respectively. 
The improvements are most pronounced in tasks requiring finer-grained understanding (Tasks B and C), while it does not show particular benefit for more coarse-grained classification tasks.

\begin{figure}[t]
  \centering
  \includegraphics[width=\linewidth]
  {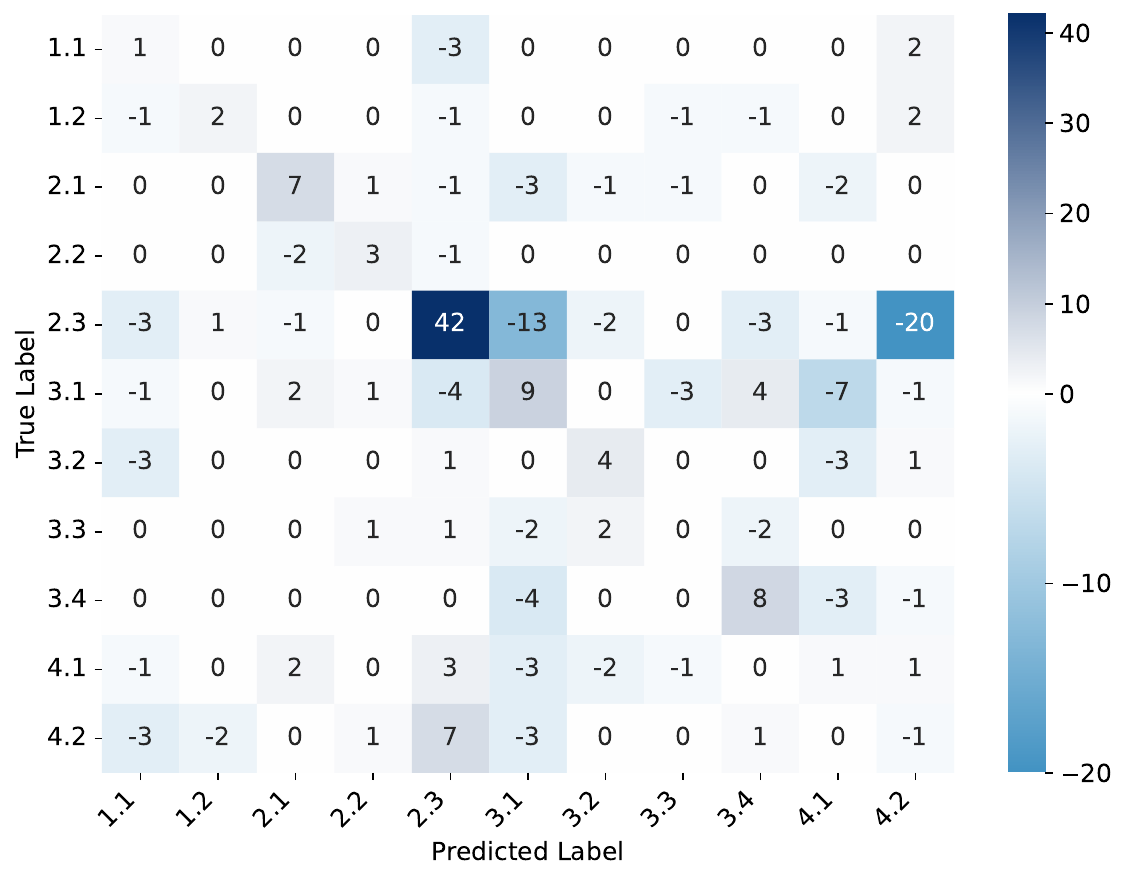}
  \caption{Difference confusion matrix for Task C (with Definition-based Data Augmentation minus without Definition-based Data Augmentation)
Positive values show where Definition-based Data Augmentation (DDA) increases counts relative to the baseline, negative values show reductions. Rows are true labels, columns predicted labels. \vspace{-8pt}}
  \label{fig:confusion-matrices}
\end{figure}

\subsection{Error Analysis}
The confusion matrix in Figure~\ref{fig:confusion-matrices} highlights the improvements achieved through Definition-based Data Augmentation (DDA), particularly benefiting underrepresented and fine-grained categories. The matrix reports the difference between the DDA model and the baseline: positive values along the diagonal indicate improved correct predictions, while negative values off the diagonal signal reduced confusion between categories. 
Compared to the baseline, DDA significantly enhances recall for classes with limited instances, such as \textit{2.3 Dehumanising attacks \& overt sexual objectification} and \textit{3.4 Condescending explanations or unwelcome advice}, where the number of correctly predicted examples increased by 42 and 8, respectively.
DDA also reduces cross-category confusion between semantically overlapping classes, such as \textit{3.1 Casual use of gendered slurs, profanities, and insults} and \textit{3.2 Immutable gender differences and gender stereotypes}. Off-diagonal errors for these two classes dropped from 48 in the baseline to 35 after applying DDA, representing an approximate 27\% improvement in alignment. 
Overall, the improvements observed with DDA are most pronounced in fine-grained classification tasks, where class boundaries are subtle and overlap is frequent.


\section{Conclusion}
In this study, we addressed the challenges of data sparsity and nuanced interpretation in fine-grained sexism detection tasks. We introduced two targeted data augmentation techniques: Definition-based Data Augmentation (DDA), which generates semantically aligned examples by leveraging explicit category definitions, and Contextual Semantic Expansion (CSE), a prompt-based method that enriches systematically misclassified examples with task-relevant contextual features. Additionally, we demonstrated the utility of multi-perspective model aggregation through the Mistral-7B Fallback Ensemble (M7-FE), which improves prediction reliability in multi-class classification by resolving ties among fine-tuned models.
Our experimental evaluation on the EDOS dataset shows that the proposed methods achieve state-of-the-art performance across all tasks. 

\section{Limitations}
First, Definition-based Data Augmentation (DDA) and Contextual Semantic Expansion (CSE) rely on prompt engineering and large language models (LLMs), which may introduce biases stemming from the models’ pretraining data. Ensuring that these synthetic examples are free of unintended biases or artifacts is an ongoing challenge that requires further investigation.
Additionally, our current evaluation focuses on the EDOS dataset as it provides detailed annotator information, it is well-curated but limited to English-language content. The generalizability of our methods to multilingual and low-resource datasets remains an open question that we aim to address in future work.
Second, although the Mistral-7B Fallback Ensemble (M7-FE) effectively resolves tie-breaking scenarios and improves reliability, it is not intended as a novel ensemble method. Exploring alternative ensemble strategies, such as weighted voting, confidence-based aggregation, or advanced meta-ensemble techniques, may further enhance performance and robustness in multi-class settings.

\section*{Acknowledgements}
Sahrish Khan, Gabriele Pergola, and Arshad Jhumka, were partially supported by the Police Science, Technology, Analysis, and Research (STAR) Fund 2022–23 and 2025-26, funded by the National Police Chiefs' Council (NPCC), in collaboration with the Forensic Capability Network (FCN). 
Gabriele Pergola was partially supported by the ESRC-funded project \textit{Digitising Identity: Navigating the Digital Immigration System and Migrant Experiences}, as part of the Digital Good Network. 
This work was conducted on the Sulis Tier-2 HPC platform hosted by the Scientific Computing Research Technology Platform at the University of Warwick. Sulis is funded by EPSRC Grant EP/T022108/1 and the HPC Midlands+ consortium.

\bibliography{custom,anthology}

\newpage

\appendix

\setcounter{table}{0}
\renewcommand{\thetable}{A\arabic{table}}
\setcounter{figure}{0}
\renewcommand{\thefigure}{A\arabic{figure}}

\clearpage
\section{Prompt Structures for Data Augmentation}
\label{appendix-baseline-prompt}

\begin{figure}[htbp]
 \centering
\includegraphics[width=\linewidth]{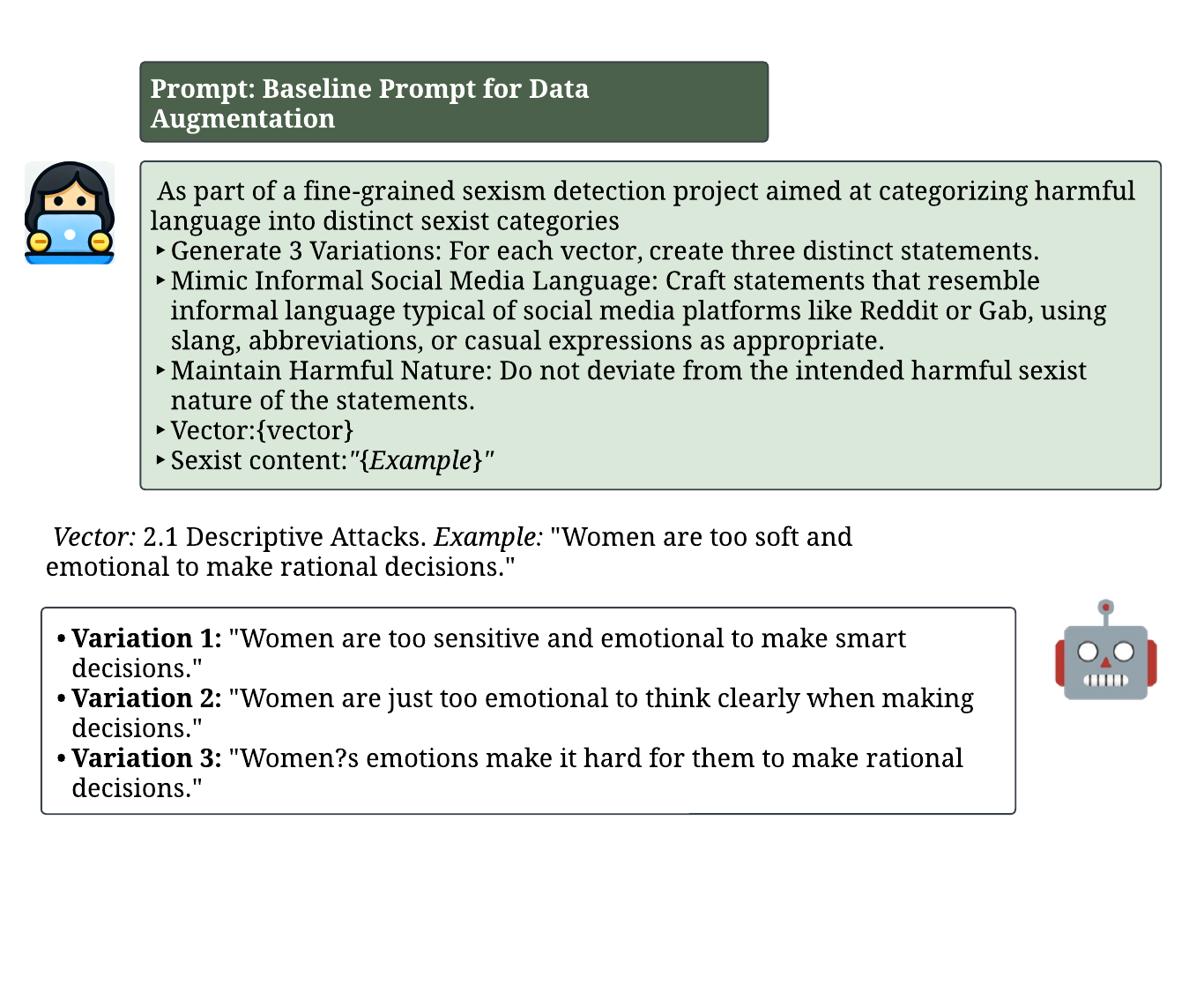}
\caption{Baseline Prompt without DDA}
\label{fig-baseline-prompt}
\end{figure}

The Baseline Prompt employed for data augmentation serves as a starting point to generate multiple variations of sexist statements without specific alignment to category definitions. Figure~\ref{fig-baseline-prompt} highlights its limitations, as the baseline lacks the ability to distinguish overlapping categories effectively. This demonstrates the need for advanced methods like DDA, which better capture the complexity of nuanced sexist language.

The section describes the prompts used for generating synthetic data to augment the dataset, focusing on the Baseline Prompt and the Definition-based Data Augmentation (DDA) Prompt. Figure~\ref{fig:dda-prompt-1} compares these prompts, highlighting how the DDA prompt incorporates nuanced definitions to align synthetic examples closely with specific categories, in contrast to the baseline prompt's generic approach. By using the DDA prompt, the generated data better captures subtle distinctions in sexist language, such as those in "2.1 Descriptive Attacks" or "3.4 Condescending Explanations," ultimately enhancing the model's ability to handle overlapping categories.

\section{Hyperparameters Search}
\label{appendix-Best-Hyperparameters-per-Model}
This section summarizes the optimal hyperparameters used for fine-tuning the models resulting from an optimised grid-search approach. Table~\ref{tab:Hyperparameters} lists parameters such as learning rates, batch sizes, and weight decay values for each model. 
\begin{table*}[htb]
\centering
\small
\begin{tabular}{lcccc}
\toprule
\textbf{Hyperparameter} & \textbf{RoBERTa-Large} & \textbf{DeBERTa-V3-Large} & \textbf{Mistral-7b} & \textbf{DTFN} \\
\midrule
Number of Epochs & 30 & 30 & 10 & 30 \\
Learning Rate & $6 \times 10^{-6}$ & $6 \times 10^{-6}$ & $1 \times 10^{-4}$ & $6 \times 10^{-6}$ \\
Batch Size & 16 & 16 & 16 & 4 \\
Weight Decay & $5 \times 10^{-3}$ & $5 \times 10^{-3}$ & $5 \times 10^{-3}$ & $5 \times 10^{-3}$ \\
\bottomrule
\end{tabular}
\caption{Best Hyperparameters per Model}
\label{tab:Hyperparameters}
\end{table*}

\section{Class-Specific Data Augmentation Strategies for Task C}
\label{appendix:class_specific_aug}

\begin{table*}[htpb]
\centering
\small
\begin{tabular}{lcc}
\hline
\textbf{Task C Classes} & \textbf{Baseline} & \textbf{DDA (3 Variations)} \\ 
\midrule
\textbf{1.1 Threats of harm} & 56 & 168 \\ 
1.2 Incitement and encouragement of harm & 254 & 762 \\ 
2.1 Descriptive attacks & 717 & 2,151 \\ 
2.2 Aggressive and emotive attacks & 673 & 2,019 \\ 
\textbf{2.3 Dehumanising attacks and overt sexual objectification} & 200 & 600 \\ 
3.1 Casual use of gendered slurs, profanities, and insults & 637 & 1,911 \\ 
3.2 Immutable gender differences and gender stereotypes & 417 & 1,251 \\ 
\textbf{3.3 Backhanded gendered compliments} & 64 & 192 \\ 
\textbf{3.4 Condescending explanations or unwelcome advice} & 47 & 141 \\ 
\textbf{4.1 Supporting mistreatment of individual women }& 75 & 225 \\ 
4.2 Supporting systemic discrimination against women as a group & 258 & 774 \\ 
\bottomrule
\end{tabular}
\caption{Count of Entries Across Vectors in Task C: Baseline and DDA (Definition-based Data Augmentation) Figures}
\label{tab:dda-data}
\end{table*}

\begin{table*}[htbp]
\centering
\begin{adjustbox}{max width=\textwidth}
\begin{tabular}{@{}lccccc@{}}
\toprule
\textbf{Model} & \textbf{Pre-training} & \textbf{Task A} & \textbf{Task B} & \textbf{Task C} \\
\midrule
\rowcolor{gray!15}
\textbf{Definition-based Data Augmentation} &  & &  & \\
\midrule
\rowcolor{gray!8}
\multicolumn{5}{@{}l}{\textbf{Generated Data with DDA: 3 Variations of 5 Key Classes }} \\
\quad DeBERTa-v3-Large (epoc:9) & $\checkmark$ & 0.8659 & 0.6885 & 0.5571 \\
\quad DeBERTa-v3-Large (epoc:7)& $\checkmark$ & 0.8648 & 0.7009 & 0.5410 \\
\quad DeBERTa-v3-Large  (epoc:3) & $\checkmark$ & 0.8614 & 0.6995 & 0.5405 \\
\quad DTFN   (epoc:3) &$\checkmark$ & 0.8656 & 0.6954 & 0.5352 \\
\quad DTFN  (epoc:7) & $\checkmark$ & 0.8667 &  0.6991& 0.5385 \\
\quad DTFN (epoc:8) & $\checkmark$ & 0.8684 & 0.6946 & 0.5344 \\
\quad Mistral-7B  & $\times$ & 0.8617 & 0.6987 & 0.5120 \\
\quad + Mistral-7B Fallback Ensemble  &  Mixed & 0.8731 & \textbf{0.7277} & \textbf{0.6018} \\
\midrule
\rowcolor{gray!8}
\multicolumn{5}{@{}l}{\textbf{Generated Data with DDA: 3 Variations of 11 Classes}} \\
\quad DeBERTa-v3-Large (epoc: 3) &  $\checkmark$ & 0.8694 & \textit{0.6903} & 0.5292 \\
\quad DeBERTa-v3-Large (epoc: 9) &  $\checkmark$ & 0.8669 & 0.6837 & \textit{0.5427} \\
\quad DeBERTa-v3-Large (epoc: 7) &  $\checkmark$ & 0.8679 & \textit{0.7032 }& 0.5357\\
\quad DTFN (epoc: 3) &  $\checkmark$& 0.8653& \textit{0.6907} & \textit{0.5383} \\
\quad DTFN (epoc: 7) &  $\checkmark$& 0.8655  &  \textit{0.6915} & \textit{0.5229} \\
\quad DTFN (epoc: 8) &  $\checkmark$& 0.8677 & 0.6866 & \textit{0.5371} \\
\quad Mistral-7B  & $\times$ & 0.8670 & \textit{0.6736} & \textit{0.4848}\\
\quad + Mistral-7B Fallback Ensemble  &  Mixed & 0.8769 & 0.7095 & 0.5608 \\
\midrule
\rowcolor{gray!8}
\multicolumn{5}{@{}l}{\textbf{Generated Data with Baseline Prompts: 3 Variations of 5 Key Classes}} \\
\quad DeBERTa-v3-Large (epoc: 3) &  $\checkmark$ & 0.8567 & 0.6748 & 0.5469 \\
\quad DeBERTa-v3-Large (epoc: 7) &  $\checkmark$ & 0.8608 & 0.6919 & 0.5248 \\
\quad DTFN (epoc: 3) &  $\checkmark$& 0.8592 & 0.6851 & 0.5163\\
\quad DTFN (epoc: 7) &  $\checkmark$& 0.8656 &  0.6977 & 0.5408 \\
\quad DTFN (epoc: 8) &  $\checkmark$& 0.8677 & 0.8674 & 0.5471 \\
\quad Mistral-7B  & $\times$ & 0.8667 & 0.6799 & 0.4909 \\
\quad + Mistral-7B Fallback Ensemble  &  Mixed & \textbf{0.8783} & 0.7049 & 0.5601 \\
\midrule
\rowcolor{gray!8}
\multicolumn{5}{@{}l}{\textbf{Generated Data with Baseline Prompts: 3 Variations of 11 Classes }} \\
\quad DeBERTa-v3-Large (epoc: 3) &  $\checkmark$ & 0.8534 & 0.6771 & 0.5218 \\
\quad DeBERTa-v3-Large (epoc: 7) &  $\checkmark$ & 0.8584 & 0.6720 & 0.5248 \\
\quad DTFN (epoc: 3) &  $\checkmark$& 0.8601 & 0.6807  & 0.4969 \\
\quad DTFN (epoc: 7) &  $\checkmark$& 0.8632 &  0.6782 & 0.5100 \\
\quad DTFN (epoc: 8) &  $\checkmark$& 0.8675 & 0.6794 & 0.5196 \\
\quad Mistral-7B  & $\times$ &0.8674 & 0.6798 & 0.4919 \\
\quad + Mistral-7B Fallback Ensemble  &  Mixed & 0.8723 & 0.6977 & 0.5361 \\

\bottomrule
\end{tabular}
\end{adjustbox}
\caption{Performance of Models on Tasks A, B, and C with Various Data Augmentation Techniques: This table compares the macro F1 scores of different models across Tasks A, B, and C, using synthetic data generated from both baseline prompts and definition-based prompts. Each model's performance is reported after training on three variations of the whole dataset or five specific classes.}
\label{all-results}
\end{table*}

Table~\ref{tab:dda-data} provides a detailed breakdown of all 11 classes within the dataset. The five underrepresented classes targeted for augmentation are highlighted in bold. These categories, selected due to their low baseline representation, limited the model's ability to effectively learn patterns from the original dataset. Augmenting only these categories using DDA significantly improved classification performance for these challenging cases.

An exhaustive and detailed comparison of model performance across three tasks (A, B, and C) using synthetic data generated by both the Baseline Prompts and Definition-based Data Augmentation (DDA) Prompts, is presented in Table~\ref{all-results}. The analysis focuses on the effect of augmenting either all eleven Task C classes or five underrepresented key classes, i.e., \textit{1.1 Threats of Harm, 2.3 Dehumanizing Attacks \& Overt Sexual Objectification, 3.3 Backhanded Gendered Compliments, 3.4 Condescending Explanations or Unwelcome Advice, and 4.1 Supporting Mistreatment of Individual Women}.
The results indicate that augmenting data for just the five key classes significantly improves the model's overall performance across all 11 Task C categories. For example, the Mistral-7B Fallback Ensemble trained with DDA-generated data for five key classes achieved a macro F1 score of 0.6018 on Task C, compared to 0.5601 when using baseline prompts for the same five classes.  In contrast, generating synthetic data for all 11 classes using DDA showed only incremental improvements, such as a macro F1 score of 0.5608 on Task C, which was marginally higher than the baseline score of 0.5361. These findings emphasize that targeted augmentation of a subset of classes is not only computationally efficient but also yields better performance gains than augmenting the entire dataset.
The comparison also confirms that DDA-based prompts outperformed baseline prompts across all models and tasks. For instance, in Task A, the DDA-trained Mistral-7B Fallback Ensemble achieved 0.8731, compared to 0.8783 using baseline prompts for five classes. This illustrates the broader applicability of DDA for fine-grained classification tasks.



\begin{table*}[htb]
\centering

\begin{adjustbox}{max width=\textwidth}
\begin{tabular}{l c l c}
\toprule
\rowcolor{gray!8}
\textbf{Definition-based Data Augmentation} &  &  &  \\
\midrule
\textbf{Tasks} & \textbf{Pretrained Epoch} & \textbf{Model} & \textbf{Macro F1 Score in Test Phase} \\
\midrule
\multirow{6}{*}{\textbf{Sub-Task A}}  & 7 & DeBERTa-v3-Large & 0.8679 \\
 & 3 & DTFN & 0.8653 \\
 & N/A & Mistral-7B & 0.8670  \\
 & Mixed &  + Mistral-7B Fallback Ensemble & \textbf{0.8769}\\
 \midrule
\multirow{6}{*}{\textbf{Sub-Task B}} & 3 & DTFN & 0.6954 \\ 
 & 7 & DTFN & 0.6991 \\
 & 3 & DeBERTa-v3-Large & 0.6995 \\
 & 7 & DeBERTa-v3-Large & 0.7009 \\
 & N/A & Mistral-7B & 0.6987 \\
 & Mixed &  + Mistral-7B Fallback Ensemble & \textbf{0.7277}\\
\midrule 
\multirow{6}{*}{\textbf{Sub-Task C}} & 3 & DTFN & 0.5352 \\ 
 & 7 & DTFN & 0.5385 \\
 & 8 & DTFN & 0.5344 \\
 & 9 & DeBERTa-v3-Large & 0.5571 \\
 & N/A & Mistral-7B & 0.5120 \\
 & Mixed &  + Mistral-7B Fallback Ensemble & \textbf{0.6018 }\\
 & Mixed &  + DTFN Fallback Ensemble & 0.5895\\
 & Mixed &  + DeBERTa-v3-Large Fallback Ensemble & 0.5877\\
\bottomrule
\end{tabular}
\end{adjustbox}
\caption{Macro F1 Score Comparison of Different Models Used as Fallback in the Ensemble for Task B and Task C with \textbf{DDA}. Mistral-7B consistently outperforms other models in both tasks, justifying its selection as the fallback model.}

\label{tab:task3_models}
\end{table*}


\begin{table*}[htbp]
\centering

\begin{adjustbox}{max width=\textwidth}
\begin{tabular}{l c l c}
\toprule
\rowcolor{gray!8}
\textbf{Contextual Semantic Expansion} &  &  &  \\
\midrule
\textbf{Tasks} & \textbf{Pretrained Epoch} & \textbf{Model} & \textbf{Macro F1 Score in Test Phase} \\
\midrule
\multirow{6}{*}{\textbf{Sub-Task A}}  & 7 & DTFN & 0.8733 \\
 & 8 & DTFN & 0.8707 \\
 & N/A & Mistral-7B & 0.8625 \\
 & Mixed &  + Mistral-7B Fallback Ensemble & \textbf{0.8818 }\\
 \midrule
\multirow{6}{*}{\textbf{Sub-Task B}} & 3 & DTFN & 0.6911 \\ 
 & 7 & DTFN & 0.6922 \\
 & 3 & DeBERTa-v3-Large & 0.6891 \\
 & 7 & DeBERTa-v3-Large & 0.6861 \\
 & N/A & Mistral-7B & 0.6891 \\
 & Mixed &  + Mistral-7B Fallback Ensemble & \textbf{0.7243}\\
\midrule 
\multirow{6}{*}{\textbf{Sub-Task C}} & 3 & DTFN & 0.5303 \\ 
 & 7 & DTFN & 0.5310 \\
 & 3 & DeBERTa-v3-Large & 0.5132 \\
 & 7 & DeBERTa-v3-Large & 0.5332\\
 & N/A & Mistral-7B & 0.5062 \\
 & Mixed &  + Mistral-7B Fallback Ensemble & \textbf{0.5639 }\\
\bottomrule
\end{tabular}
\end{adjustbox}
\caption{Macro F1 Score Comparison of Different Models Used as Fallback in the Ensemble for Task B and Task C with \textbf{CSE}. Mistral-7B Fallback Ensemble consistently outperforms other models in both tasks, justifying its selection as the fallback model.}

\label{tab:individual-model-results-cse}
\end{table*}


\subsection{Analysis of Annotator Agreement and Disagreement Across Task C Categories}
\label{Human-Annotator-Analysis}

Table~\ref{tab:agreement_stats-percentage} provides detailed statistics on annotator agreement, revealing key trends and challenges in the labeling process for complex sexism categories. Full agreement, where all three annotators assign the same label, is notably rare across categories, with an average agreement of $15\%$. The highest full agreement is observed in 3.1 Casual use of slurs, profanities, and insults at $28.9\%$, while categories such as 3.3 Backhanded gendered compliments and 3.4 Condescending explanations or unwelcome advice show $0\%$ and $2.9\%$ full agreement, respectively.  Full agreement indicates instances where the sexist content is explicit and unambiguous, facilitating consistent labeling. However, its rarity underscores the inherent subjectivity and complexity of interpreting subtle and context dependent language. 

Partial disagreement, where at least two annotators agree on one label while the third annotator selects a different label, is the most common outcome, dominating across categories like 2.1 Descriptive attacks $50\%$ and 2.2 Aggressive and emotive attacks $49.8\%$. This trend highlights the difficulty in differentiating overlapping or closely related categories, such as distinguishing between targeted and systemic harm.  

Full disagreement, where all three annotators assign different labels, is also prevalent in complex categories like 3.4 Condescending explanations or unwelcome advice $85.7\%$, reflecting the subjective interpretations of implicit biases or condescending tones. These findings underscore the importance of refining category definitions and providing contextual guidance to mitigate ambiguity and enhance agreement.

\begin{table*}[htbp]
\centering
\small
\setlength{\tabcolsep}{2.5pt} 
\renewcommand{\arraystretch}{1.1} 
\begin{tabular}{lcccc}
\toprule
\rowcolor{gray!10}
\textbf{Task C Vector} & \textbf{Full} & \textbf{Partial} & \textbf{Full} &\\
\rowcolor{gray!10}
 & \textbf{Agreement} & \textbf{Disagreement} & \textbf{Disagreement} & \textbf{Number of Instances}\\ 
\midrule
1.1 Threats of harm & 16.2\% & 40\% & 43.7\% & 80 \\ 
1.2 Incitement and encouragement of harm & 19.8\% & 41\% & 39.1\% & 363 \\ 
2.1 Descriptive attacks & 17.9\% & 50\% & 32.1\% & 1024 \\ 
2.2 Aggressive and emotive attacks & 16.5\% & 49.8\% & 33.6\% & 961\\ 
2.3 Dehumanising attacks \& sexual objectification & 13.9\% & 40.8\% & 40.2\% & 286 \\ 
3.1 Casual use of slurs, profanities, and insults & 28.9\% & 49.5\% & 21.5\% & 910 \\ 
3.2 Immutable gender differences \& stereotypes & 6.8\% & 31.1\% & 62.1\% & 596 \\ 
3.3 Backhanded gendered compliments & 0\% & 20.9\% & 79.1\% &  91 \\ 
3.4 Condescending explanations or unwelcome advice & 2.9\% & 39.7\% & 57.4\% & 68 \\ 
4.1 Supporting mistreatment of individual women & 8.4\% & 41.1\% & 50.5\% & 107 \\ 
4.2 Supporting systemic discrimination & 9.8\% & 40.2\% & 50\% & 368 \\
\bottomrule
\end{tabular}
\caption{Human Annotator Agreement and Disagreement Statistics Across Task C Categories: This table highlights the instances of full agreement, partial disagreement, and full disagreement among annotators for each sexism subcategory, demonstrating challenges in achieving consensus. These statistics reflect the entire dataset, including examples used in training, evaluation, and the test set, showcasing the complexities involved in fine-grained sexism classification and the potential difficulties for models in differentiating between closely related categories.}
\label{tab:agreement_stats-percentage}
\end{table*}

\subsection{Analysis of Ensemble Agreement}

The agreement and disagreement trends among ensemble models are reported in Table~\ref{tab:llm-test-agreement_stats}. Categories such as ``1.1 Threats of harm” (62.5\% full agreement) and ``1.2 Incitement and encouragement of harm” (58.9\%) exhibit high agreement rates, indicating that explicit forms of sexism with clear linguistic markers are easier for models to classify consistently. In contrast, nuanced and conceptually overlapping categories like ``3.3 Backhanded gendered compliments” (27.8\% full agreement) and ``3.4 Condescending explanations or unwelcome advice” (14.3\%) demonstrate high partial disagreement rates (72.2\% and 85.7\%, respectively), underscoring the difficulty in capturing implicit and subjective forms of sexism. Furthermore, low-resource categories with fewer instances, such as ``3.3” and ``3.4,” face exacerbated challenges, highlighting the need for targeted data augmentation strategies like DDA to enhance model performance. Notably, the absence of full disagreement in nearly all categories suggests that ensemble learning provides robust predictions, leveraging complementary strengths of individual models to mitigate classification errors. These findings underscore the importance of refining category definitions, improving contextual augmentation, and exploring interpretability techniques to address persistent challenges in fine-grained sexism detection.

\begin{table*}[htbp]
\centering
\small
\setlength{\tabcolsep}{2.5pt} 
\renewcommand{\arraystretch}{1.1} 
\begin{tabular}{lcccc}
\toprule
\rowcolor{gray!10}
\textbf{Task C Vector} & \textbf{Full} & \textbf{Partial} & \textbf{Full} & \textbf{Number of } \\
\rowcolor{gray!10}
 & \textbf{Agreement} & \textbf{Disagreement} & \textbf{Disagreement} & \textbf{Instances} \\ 
\midrule
1.1 Threats of harm & 62.5\% & 37.5\% & 0 & 16 \\ 
1.2 Incitement and encouragement of harm & 58.9\% & 41.1\% & 0 & 73 \\ 
2.1 Descriptive attacks & 43.4\% & 56.1\% & 0.5\% & 205 \\ 
2.2 Aggressive and emotive attacks & 42.2\% & 57.8\% & 0\% & 192 \\ 
2.3 Dehumanising attacks \& sexual objectification & 43.9\% & 56.1\% & 0\% & 57 \\ 
3.1 Casual use of slurs, profanities, and insults & 52.2\% & 47.8\% & 0\% & 182\\ 
3.2 Immutable gender differences \& stereotypes & 47\% & 53\% & 0\% & 119 \\ 
3.3 Backhanded gendered compliments & 27.8\% & 72.2\% & 0\% & 18 \\ 
3.4 Condescending explanations or unwelcome advice & 14.3\% & 85.7\% & 0\% & 14 \\ 
4.1 Supporting mistreatment of individual women & 33.3\% & 66.7\% & 0\% & 21 \\ 
4.2 Supporting systemic discrimination & 48\% & 52\% & 0\% & 73\\ 
\bottomrule
\end{tabular}
\caption{Prediction Agreement and Disagreement Among Five Models in Ensemble: This table illustrates the counts of full agreement, partial disagreement, and full disagreement among model predictions for Task C categories of the Test Set, showcasing the distribution of model votes in the ensemble.}
\label{tab:llm-test-agreement_stats}
\end{table*}

\begin{table*}[htbp]
\centering
\scalebox{0.85} {
\begin{tabular}{lcc}
\toprule
\rowcolor{gray!15}
\textbf{Aggregated Label} & \textbf{Count of Discrepant Examples} & \textbf{Split} \\  \midrule
1.1 threats of harm & 3 & dev \\ 
1.2 incitement and encouragement of harm & 3 & dev \\ 
2.1 descriptive attacks & 5 & dev \\ 
2.2 aggressive and emotive attacks & 11 & dev \\ 
2.3 dehumanising attacks \& overt sexual objectification & 3 & dev \\ 
3.1 casual use of gendered slurs, profanities, etc. & 5 & dev \\ 
3.2 immutable gender differences and gender stereotypes & 18 & dev \\ 
3.3 backhanded gendered compliments & 5 & dev \\ 
3.4 condescending explanations or unwelcome advice & 1 & dev \\ 
4.1 supporting mistreatment of individual women & 1 & dev \\ 
4.2 supporting systemic discrimination against women as a group & 5 & dev \\ \midrule 
1.1 threats of harm & 2 & test \\ 
1.2 incitement and encouragement of harm & 9 & test \\ 
2.1 descriptive attacks & 3 & test \\ 
2.2 aggressive and emotive attacks & 23 & test \\ 
2.3 dehumanising attacks \& overt sexual objectification & 8 & test \\ 
3.1 casual use of gendered slurs, profanities, etc. & 4 & test \\ 
3.2 immutable gender differences and gender stereotypes & 24 & test \\ 
3.3 backhanded gendered compliments & 7 & test \\ 
3.4 condescending explanations or unwelcome advice & 2 & test \\ 
4.1 supporting mistreatment of individual women & 5 & test \\ 
4.2 supporting systemic discrimination against women as a group & 7 & test \\ 
none-sexist & 4 & test \\  \midrule
1.1 threats of harm & 9 & train \\ 
1.2 incitement and encouragement of harm & 25 & train \\ 
2.1 descriptive attacks & 41 & train \\ 
2.2 aggressive and emotive attacks & 50 & train \\ 
2.3 dehumanising attacks \& overt sexual objectification & 12 & train \\ 
3.1 casual use of gendered slurs, profanities, etc. & 28 & train \\ 
3.2 immutable gender differences and gender stereotypes & 94 & train \\ 
3.3 backhanded gendered compliments & 22 & train \\ 
3.4 condescending explanations or unwelcome advice & 7 & train \\ 
4.1 supporting mistreatment of individual women & 7 & train \\ 
4.2 supporting systemic discrimination against women as a group & 27 & train \\ 
none-sexist & 10 & train \\ \midrule
\end{tabular} 
}
\caption{Aggregated Split and Label Counts: Number of discrepant examples in each split where the aggregated label differs from the individual annotator labels.}
\label{tab:aggregated_split_counts}
\end{table*}

\section{Definition-based Data Augmentation (DDA) Prompt}
\label{appendix:sexism_definition}
This appendix reports the detailed category definitions adopted from the EDOS dataset \cite{edos_dataset}. These definitions, shown in Table~\ref{definition}, are employed as part of the Definition-based Data Augmentation (DDA) Prompt, as shown in the example of Figure~\ref{fig:dda-prompt-1}

\begin{table*}[htbp]
\centering
\small

\renewcommand{\arraystretch}{1.6} 
\begin{tabular}{|p{3.6cm}|p{2.6cm}|p{8.4cm}|} 
\hline
\rowcolor[HTML]{C9DAF8} 
\textbf{Category} & \textbf{Vector} & \textbf{Definition} \\
\hline
\multirow{2}{*}{\cellcolor[HTML]{E7E7E7} \makecell[l]{1. Threats, plans to\\harm and incitement}}
 & \cellcolor[HTML]{FCE5CD} 1.1 Threats of harm & \cellcolor[HTML]{FCE5CD} Expressing intent, willingness, or desire to harm an individual woman or group of women. This could include physical, sexual, emotional, or privacy-based forms of harm. \\
\cline{2-3}
 \cellcolor[HTML]{E7E7E7}& \cellcolor[HTML]{FCE5CD} 1.2 Incitement and encouragement of harm & \cellcolor[HTML]{FCE5CD} Inciting or encouraging an individual, group, or general audience to harm a woman or group of women. It includes language where the author seeks to rationalize and/or justify harming women to another person. \\
\hline
\multirow{3}{*}{\cellcolor[HTML]{FAD7A0} 2. Derogation} 
 & \cellcolor[HTML]{FFF2CC} 2.1 Descriptive attacks & \cellcolor[HTML]{FFF2CC} Characterizing or describing women in a derogatory manner. This could include negative generalizations about women’s abilities, appearance, sexual behavior, intellect, character, or morals. \\
\cline{2-3}
 \cellcolor[HTML]{FAD7A0} & \cellcolor[HTML]{FFF2CC} 2.2 Aggressive and emotive attacks & \cellcolor[HTML]{FFF2CC} Expressing strong negative sentiment against women, such as disgust or hatred. This can be through direct description of the speaker’s subjective emotions, baseless accusations, or the use of gendered slurs, gender-based profanities, and gender-based insults. \\
\cline{2-3}
 \cellcolor[HTML]{FAD7A0} & \cellcolor[HTML]{FFF2CC} 2.3 Dehumanizing attacks and overt sexual objectification & \cellcolor[HTML]{FFF2CC} Derogating women by comparing them to non-human entities such as vermin, disease, or refuse, or overtly reducing them to sexual objects. \\
\hline
\multirow{4}{*}{\cellcolor[HTML]{D0ECE7} 3. Animosity} 
 & \cellcolor[HTML]{D9EAD3} 3.1 Casual use of gendered slurs, profanities, and insults & \cellcolor[HTML]{D9EAD3} Using gendered slurs, gender-based profanities, and insults, but not to intentionally attack women. Only terms that traditionally describe women are in scope (e.g., ‘b*tch’, ‘sl*t’). \\
\cline{2-3}
 \cellcolor[HTML]{D0ECE7}  & \cellcolor[HTML]{D9EAD3} 3.2 Immutable gender differences and gender stereotypes & \cellcolor[HTML]{D9EAD3} Asserting immutable, natural, or otherwise essential differences between men and women. In some cases, this could be in the form of using women’s traits to attack men. Most sexist jokes will fall into this category. \\
\cline{2-3}
\cellcolor[HTML]{D0ECE7}  & \cellcolor[HTML]{D9EAD3} 3.3 Backhanded gendered compliments & \cellcolor[HTML]{D9EAD3} Ostensibly complimenting women, but actually belittling or implying their inferiority. This could include reduction of women’s value to their attractiveness or implication that women are innately frail, helpless, or weak. \\
\cline{2-3}
\cellcolor[HTML]{D0ECE7}  & \cellcolor[HTML]{D9EAD3} 3.4 Condescending explanations or unwelcome advice & \cellcolor[HTML]{D9EAD3} Offering unsolicited or patronizing advice to women on topics and issues they know more about (known as ‘mansplaining’). \\
\hline
\multirow{2}{*}{\cellcolor[HTML]{F4CCCC} 4. Prejudiced Discussion} 
 & \cellcolor[HTML]{EAD1DC} 4.1 Supporting mistreatment of individual women & \cellcolor[HTML]{EAD1DC} Expressing support for mistreatment of women as individuals. Support can be shown by denying, understating, or seeking to justify such mistreatment. \\
\cline{2-3}
 \cellcolor[HTML]{F4CCCC} & \cellcolor[HTML]{EAD1DC} 4.2 Supporting systemic discrimination against women as a group & \cellcolor[HTML]{EAD1DC} Expressing support for systemic discrimination of women as a group. Support can be shown by denying, understating, or seeking to justify such discrimination. \\
\hline
\end{tabular}
\caption{Taxonomy of Sexism Categories for Task B and Fine-Grained Vectors in Task C: This table presents the detailed definitions for each sexism category and corresponding fine-grained vectors. The categories span threats, derogation, animosity, and prejudiced discussion, providing a comprehensive framework for distinguishing between nuanced forms of sexism.}
\label{definition}
\end{table*}

\end{document}